\documentclass[journal]{IEEEtran}

\usepackage{cite}
\usepackage{graphicx}
\usepackage{array}
\usepackage{comment}
\usepackage{amsmath,amssymb,bm}

\usepackage{graphicx}
\usepackage{subfigure}
\usepackage{multirow}
\usepackage{xspace}
\usepackage{amsmath}
\usepackage{amssymb}
\usepackage{soul}
\usepackage{color}
\usepackage[colorlinks,
            linkcolor=black,
            anchorcolor=black,  
            citecolor=black,       
            ]{hyperref}

\makeatletter
\DeclareRobustCommand\onedot{\futurelet\@let@token\@onedot}
\def\@onedot{\ifx\@let@token.\else.\null\fi\xspace}

\def\eg{\emph{e.g.,}\xspace}
\def\ie{\emph{i.e.,}\xspace} 

\def\etc{\emph{etc}\onedot}
\def\etal{\emph{et al}\onedot}

\newcommand{\PEB}{Pose Estimation branch\xspace}
\newcommand{\RB}{ReID branch\xspace}
\newcommand{\MSP}{Modality-specific module\xspace}
\newcommand{\MSH}{Modality-shared module\xspace}
\newcommand{\HFC}{Hierarchical Feature Constraint\xspace}
\newcommand{\RM}{Refinement Module\xspace}

\begin{document}

\begin{titlepage}
This work has been submitted to the IEEE for possible publication. Copyright may be transferred
without notice, after which this version may no longer be accessible.
\end{titlepage}

\title{On Exploring Pose Estimation as an Auxiliary Learning Task for Visible-Infrared Person Re-identification}

\author{Yunqi~Miao,
        Nianchang~Huang,
        Xiao~Ma,
        Qiang~Zhang,
   		and~Jungong~Han
\thanks{Y. Miao and X.Ma are with the Warwick Manufacturing Group (WMG), University of Warwick, Coventry, CV4 7AL, United Kingdom (e-mail: Yunqi.Miao.1@warwick.ac.uk; X.Ma@warwick.ac.uk).}
\thanks{N. Huang and Q.Zhang are with the School of Mechano-Electronic Engineering, Xidian University, Xi'an, 710071, China (email: nchuang@stu.xidian.edu.cn; qzhang@xidian.edu.cn).}
\thanks{J. Han is with the Department of Computer Science, Aberystwyth University, Wales, SY23 3FL, United Kingdom (e-mail: jungonghan77@gmail.com).}
}

\markboth{Journal of \LaTeX\ Class Files,~Vol.~14, No.~8, August~2015}%
{Shell \MakeLowercase{\textit{et al.}}: Bare Demo of IEEEtran.cls for IEEE Journals}
%


\maketitle
 
\begin{abstract}
Visible-infrared person re-identification (VI-ReID) has been challenging due to the existence of large discrepancies between visible and infrared modalities. Most pioneering approaches reduce intra-class variations and inter-modality discrepancies by learning modality-shared and ID-related features. However, an explicit modality-shared cue, \ie body keypoints, has not been fully exploited in VI-ReID. Additionally, existing feature learning paradigms imposed constraints on either global features or partitioned feature stripes, which neglect the prediction consistency of global and part features. To address the above problems, we exploit Pose Estimation as an auxiliary learning task to assist the VI-ReID task in an end-to-end framework. By jointly training these two tasks in a mutually beneficial manner, our model learns higher quality modality-shared and ID-related features. On top of it, the learnings of global features and local features are seamlessly synchronized by \HFC (HFC), where the former supervises the latter using the knowledge distillation strategy. Experimental results on two benchmark VI-ReID datasets show that the proposed method consistently improves state-of-the-art methods by significant margins. Specifically, our method achieves nearly 20$\%$ mAP improvements against the state-of-the-art method on the RegDB dataset. Our intriguing findings highlight the usage of auxiliary task learning in VI-ReID. Our source code is available at \url{https://github.com/yoqim/Pose\_VIReID}.
\end{abstract}


%
\IEEEpeerreviewmaketitle

\section{Introduction}  
\label{Intro}
\IEEEPARstart{P}erson re-identification (ReID) aims at retrieving the same identity across multiple disjoint cameras, which has gained much attention from the recent computer vision community \cite{leng2019survey, ye2020deep, si2017spatial, zheng2018subspace, lei2021tdr}. Most existing ReID methods focus on the matching between visible images, which are generally collected under good illumination conditions \cite{xiao2018nips, chen2019iccv, guo2019iccv, zheng2019cvpr, lv2020nips}. However, those systems seem impractical because visible images cannot provide sufficient discriminatory information in poor lighting environments, \eg at night. To this end, Visible-Infrared person Re-identification (VI-ReID) emerges as an alternative to performing the retrieval between visible (RGB) images and infrared (IR) counterparts, thus enabling the day-to-night person re-identification.

\begin{figure*}
	\label{fig:1}
	\centering
	\setlength{\abovecaptionskip}{0.1cm}
	\includegraphics[width=0.95\textwidth]{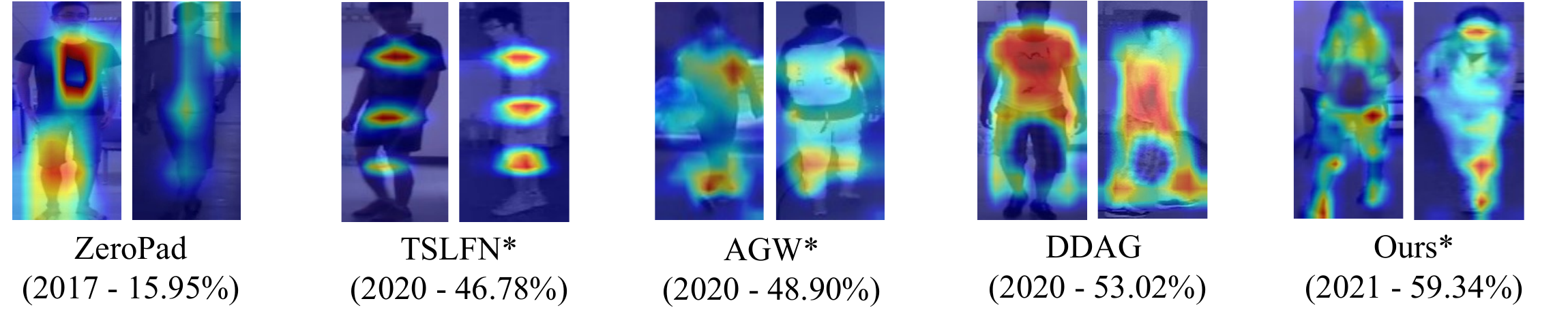}
	\caption{Visualization of features derived by ZeroPad \cite{wu2017rgb}, TSLFN \cite{zhu2020hetero}, AGW \cite{ye2020deep}, DDAG \cite{luo2020dynamic}, and our method on SYSU-MM01 dataset. For each method, the RGB image of an identity is shown on the left and the IR image is on the right. The publication year and mAP score ($\%$) at \textit{All-search} setting of each method are also reported. For a fair comparison, the mAP scores of TSLFN, AGW and our method are given by models trained merely with vanilla identity loss and triplet loss.}
\end{figure*}
However, VI-ReID is a challenging problem due to large intra-class variations and modality discrepancies across different cameras. The former refers to identity's appearance differences within a modality caused by poses, clothes, viewpoints, \etc. While, the latter denotes intrinsic differences between visible and infrared images caused by the spectrum of cameras. To reduce both discrepancies, the central research question in this field has always been seeking better ways to extract discriminative features for identity retrieval, which are ID-related and modality-invariant \cite{wu2017rgb,ye2020deep,zhu2020hetero,luo2020dynamic}. 

Despite the vigorous development of this field, we realize that most algorithms extract identity features in a heuristics manner with NO common view of what sort of features are specifically helpful for VI-ReID. To tackle this problem, we visualize the features extracted by several representative methods in Fig.~1, aiming to investigate how visual feature extractions evolved over the years to improve the performance of VI-ReID systems. Concretely, ZeroPad \cite{wu2017rgb}, as the first work in VI-ReID, extracts features from random regions in an image. Additionally, for a specific identity, the features derived from two modalities share NO commonality. As a result, the mean Average Precision (mAP) score is far from satisfactory. Later, TSLFN \cite{zhu2020hetero} horizontally partitions the backbone feature maps (global) into several stripes (local) and employs local-level constraints on each of them. Clearly, features derived by TSLFN cover more parts of the human body, compared to the global-level constraint-based method, \eg ZeroPad. That might be the reason why its performance is significantly superior to that of ZeroPad. In the meantime, a baseline for VI-ReID (AGW \cite{ye2020deep}) inserts non-local attention blocks during the feature extraction, which enforces the features to be extracted from identity's body instead of backgrounds. It reveals that \textit{attention-aware} features help to increase performance. Recently, a state-of-the-art approach (DDAG \cite{luo2020dynamic}) integrates both local-level and global-level constraints into an end-to-end framework. Compared to previous works, the features derived by DDAG are more fine-grained, which are not only shared by two modalities but distinguishable for different identities. Based on the above observations, a conclusion can be drawn: \textit{as more features from the human region (modality-shared) and attentive body part (ID-related) features are extracted, the retrieval performance improves consistently.} Benefiting from this conclusion, our algorithm extracts more features from body skeleton joints, which are not only ID-related but immune to modality changes. Therefore, we achieve over 10$\%$ mAP improvements against DDAG on the challenging SYSU-MM01 dataset at \textit{All-search} setting.

Having depicted the visual features that are conducive to VI-ReID, the next question is: how can we extract them effectively. It is noted that body skeleton points are explicit modality-shared cues and the features describing certain skeleton points are ID-related. In light of this, in the paper, we aim to facilitate the extraction of discriminative features for identity retrieval with the aid of the pose estimation task. However, making effective use of the pose information for cross-modality ReID does not seem easy, though it has recently appeared to be exploited in some single-modality ReID works \cite{zhao2017spindle,su2017pose,zheng2019pose,zhao2017deeply,suh2018part,miao2019pose}, where only visible images are involved. Earlier methods \cite{zhao2017spindle,su2017pose,zheng2019pose} utilize detected body joints to segment \cite{zhao2017spindle} or align \cite{su2017pose,zheng2019pose} body regions in order to cope with human pose changes. After the calibration of body parts, different body parts need to be stitched, which usually yields unrealistic transformed visual features. If we move to the cross-modality setting, the transformed errors would be further magnified due to the huge discrepancy between the two modalities. Alternatively, another group of methods employ the body keypoints information to refine ID-related feature maps either by means of highlighting discriminative body regions \cite{zhao2017deeply,miao2019pose} or complementing human appearance features \cite{suh2018part}. Although pose-assisted features are proved to improve the feature discriminability under the single-modality setting, they are rigidly based on the outputs of off-the-shelf pose estimators. However, such a blind trust in the pre-trained pose estimators will lead to a poor re-identification performance if the gap between the source domain and the target domain of the pose estimator is huge. Therefore, employing pose information in the VI-ReID task is extremely challenging due to the massive gap between visible images (source domain) and infrared images (target domain). To highlight the problem, we extend two state-of-the-art pose-guided single-modality person ReID methods \cite{suh2018part,miao2019pose} to the VI-ReID task and the results turn out that the best performance in the mAP score \cite{suh2018part} only reaches 42.19$\%$, which is far from satisfaction due to the inadequate usage of the pose information (more results and comparisons are provided in Section~\ref{sec:CRR}).

To solve the above problems, we propose a two-stream VI-ReID framework, where modality-shared and ID-related features for identity retrieval are extracted by means of learning an auxiliary task (pose estimation) and the main task (person ReID) jointly. Unlike previous works, which rely dramatically on the off-the-shelf pose estimators, pose features are adaptively adjusted to facilitate the ReID task in our work. Additionally, apart from ID-related constraints, an extra constraint is imposed on the pose estimation branch, ensuring that not only body skeleton points are precisely estimated but also the ID-related information are fully embedded in feature maps, \ie at both local and global levels. Despite the significant improvements obtained by the horizontal-divided feature constraints \cite{sun2018beyond} in the VI-ReID task \cite{zhu2020hetero,luo2020dynamic}, the learning of individual striped features is generally independent and its discriminability consistency with global backbone features is neglected. To this end, we propose a \HFC (HFC), in the paper, which bonds the learnings of global features and local ones via the knowledge distillation strategy. Concretely, predictions of backbone features serve as ``soft-target'' to supervise the learning of partitioned feature stripes, hence preserving the discriminability consistency of global features and local ones.

In summary, the contributions made in our work are mainly three-fold:
\begin{itemize}
	\item A novel two-stream framework for VI-ReID is proposed, where the pose estimation, for the first time, acts as an auxiliary learning task to help the ReID task in VI-ReID. To learn fine-grained pose features embedded with ID-related information, both pose and ReID constraints are imposed on the pose estimation branch.
	\item Instead of imposing feature constraints on local feature stripes only, \HFC (HFC) is proposed to ensure the discriminability consistency of global features and local ones via the knowledge distillation strategy.
	\item The proposed method performs far better than the state-of-the-art methods on two benchmark datasets: SYSU-MM01 \cite{wu2017rgb} and RegDB \cite{nguyen2017person}.
\end{itemize}

\section{Related work}
\subsection{Auxiliary tasks in person ReID}
\label{sec:RWA}
Research has shown that semantic information such as body parts, human pose \etc, can significantly facilitate the person ReID task \cite{leng2019survey,ye2020deep}. Therefore, recent works utilize auxiliary tasks to improve the performance of ReID models, which can be mainly categorized as attribute-guided methods \cite{ling2019improving,tay2019aanet}, segmentation-guided methods \cite{kalayeh2018human,song2018mask,zhu2020identity}, and pose-guided methods \cite{ge2018fd,liu2018pose,qian2018pose,zhao2017spindle,su2017pose,zheng2019pose,zhao2017deeply,suh2018part}.

Attribute information provides complementary details of identities for person ReID. Ling \etal \cite{ling2019improving} propose a multi-task learning framework, which attempts to improve the discriminability of identity features by embedding attribute information. Tay \etal \cite{tay2019aanet} enhance the identity features with attribute attention maps, where class-sensitive activation regions of various attributes, \ie clothing color, hair, gender \etc are emphasized.

Segmentation-guided methods are based on the pixel-level body parts segmentation, which improve the discriminability of identity features by masking out backgrounds \cite{song2018mask} or leverage fine-grained local features from discriminative body regions \cite{kalayeh2018human,zhu2020identity}. However, both attribute- and segmentation-assisted methods require extensive additional annotations, which are too expensive to obtain in real-world applications. Additionally, since color information is not available in infrared images, the usage of color-related attributes is restricted in VI-ReID, thereby limiting the performance improvement.

The pose estimation is employed to facilitate person ReID from mainly two aspects: 1) generating person images with various poses to augment training data \cite{ge2018fd,liu2018pose,qian2018pose}, and 2) aligning body parts \cite{zhao2017spindle,su2017pose,zheng2019pose,zhao2017deeply,suh2018part,miao2019pose}. With the help of Generative Adversarial Networks (GAN), Qian \etal \cite{qian2018pose} synthesize eight new images for an identity with a set of target canonical poses. In addition to generating visually preferable samples, Liu \etal \cite{liu2018pose} propose a guider module to ensure that the generated samples have discriminative power for ReID. Apart from image generation, pose information is employed to handle the problem of feature misalignment in ReID \cite{zhao2017spindle,su2017pose,zheng2019pose}. Zhao \etal \cite{zhao2017spindle} segment the human body into seven regions by human landmark information and combine the representations over them for identity retrieval. Zheng \etal \cite{zheng2019pose} align pedestrians to a standard pose by stitching the segmented body regions with affine transformations. However, such unnatural stitch destroys the authenticity of human and requires an elaborately designed following fusion method to fuse the local features. Instead of using detected body regions rigidly, Zhao \etal \cite{zhao2017deeply} and Miao \etal \cite{miao2019pose} employ body joint maps to refine image feature maps, where discriminative body parts for person ReID are emphasised. However, both approaches are rigidly based on the outputs of off-the-shelf pose estimators, which may generate unreliable information because of the gap between the source domain and the target domain. To avoid the blind trust in the pre-trained pose estimator, Suh \etal \cite{suh2018part} take an on-the-fly pose estimator as an individual branch to derive pose features, which are then aggregated with the appearance features from the other branch via a bilinear pooling layer. Such combined training encourages the pose branch to learn features that are beneficial to the person ReID task. While pose features are learned to adapt to the ReID task in the work, only ID-related constraints are imposed on the fused features, which ignores the quality of pose features from the individual branch. Given the above concerns, the auxiliary pose estimation task designed for single-modality ReID cannot perform well in the VI-ReID task. In other words, how to enable the pose information to facilitate the VI-ReID task has not been elaborately investigated.

Therefore, in the paper, we propose a novel pose estimation assisted framework for VI-ReID, where body skeleton points cues are learned under both pose and identity guidance, which are then employed to enhance the discriminative identity feature extraction for the cross-modality person re-identification.

\subsection{Feature constraints in VI-ReID}
To improve the discriminability of learned features, most existing works impose either global-level feature constraints on backbone convolutional features or local-level feature constraints on partitioned feature stripes. 

As a representative of global-level constraints, Ye \etal \cite{ye2018hierarchical} propose a two-stream network, which jointly optimizes modality-specific and modality-shared metrics. Based on the idea, a bi-directional top-ranking loss is then introduced in \cite{ye2019bi} to incorporate the above two constraints. Alternatively, AGW \cite{ye2020deep} presents a weighted regularized triplet loss to embed the neighboring relationship of images from two modalities in a common feature space. 

Inspired by the competitive performance of the Part-based Convolutional Baseline (PCB) model \cite{sun2018beyond} in single-modality ReID, recent VI-ReID studies start imposing local-level feature constraints on feature stripes obtained by partitioning backbone convolutional features. a kickoff work \cite{zhu2020hetero} presents a center-based loss that pulls the centers of RGB features and IR features of a given identity closer. On top of it, Liu \etal \cite{liu2020parameter} present a hetero-center triplet loss, where the feature centers of different identities are pushed away. Recently, the newly-proposed DDAG \cite{luo2020dynamic} simultaneously mines both cross-modality global-level and intra-modality part-level contextual cues, which achieves the state-of-the-art performance. Although both global-level and local-level constraints are considered in DDAG, those constraints are imposed independently without taking the discriminability consistency of them into account. 

To this end, we propose a \HFC (HFC) to bond the global feature learning with the local ones, where predictions of global features supervise the learning of part features via the knowledge distillation strategy.

\section{Proposed method}
\begin{figure*}[]
	\centering
	\includegraphics[width=\textwidth]{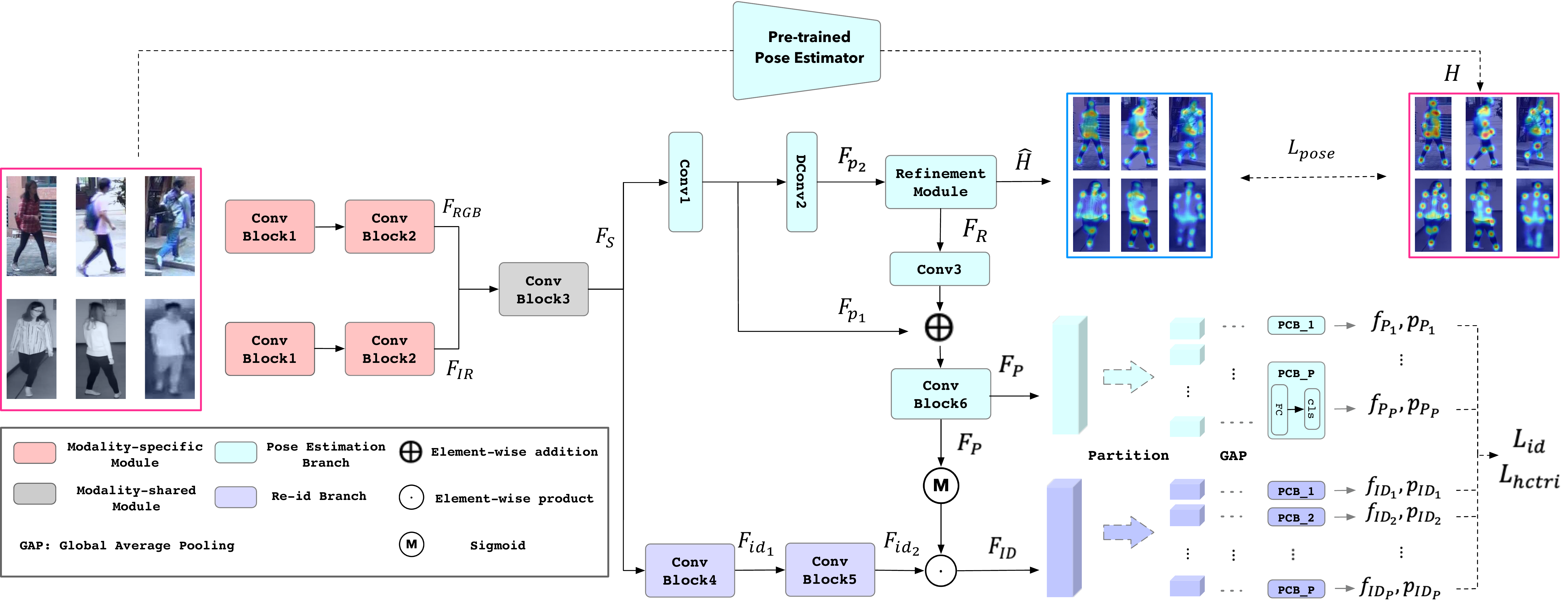}
	\caption{The framework of the proposed method.}
	\label{fig:framework}
	\vspace{0cm}
\end{figure*}

We propose a novel pose estimation assisted framework for VI-ReID, which aims to learn modality-shared and ID-related features for identity retrieval. The framework of the proposed method is shown in Fig.~\ref{fig:framework}. As can be seen, our model mainly consists of four components: \MSP, \MSH, \PEB, and \RB. Details about these components will be discussed in the following subsections.
\subsection{\MSP and \MSH}
Following previous works \cite{ye2019bi,liu2020parameter,zhu2020hetero}, ResNet50 \cite{he2016deep} is exploited as a backbone feature extractor to provide discriminative features for both pose estimation and ReID tasks. Specifically, \MSP consists of two blocks (\textit{``Conv Block1\textasciitilde2''}), which adopt the structures of shallow convolution block (\textit{layer0}) and the first res-convolution block (\textit{layer1}) of ResNet50, respectively. Note that the parameters of \MSP for RGB and IR modalities are separately updated. Then, modality-specific features of two modalities are projected into a shared feature space by the \MSH (\textit{``Conv Block3''}), which adopts the structure of the second res-convolution block (\textit{layer2}) of ResNet50.

Mathematically, given a RGB image $I_{RGB}$ and an IR image $I_{IR}$, modality-specific features $\textbf{F}_{m}$ and modality-shared features $\textbf{F}_S$ can be obtained by,
\begin{equation}
\begin{split}
\textbf{F}_m&= ConvB2(ConvB1(I_m, \theta^1_m), \theta^2_m), \quad m\in\{RGB,IR\}, \\
\textbf{F}_S&= ConvB3([\textbf{F}_{RGB}, \textbf{F}_{IR}]_b, \theta^3),
\label{equ:Fs}
\end{split}
\end{equation}
where $[\cdot,\cdot]_b$ represents the feature concatenation along the data dimension. $ConvB1(*,\theta_m^1)$, $ConvB2(*,\theta_m^2)$, and $ConvB3(*,\theta^3)$ denote the convolution blocks of \MSP and \MSH with corresponding parameters $\theta^1_m$, $\theta^2_m$ and $\theta^3$, respectively.

\subsection{\PEB}
\subsubsection{Body keypoint features extraction}
Given the observation that modality-shared features are beneficial to the VI-ReID task, a \PEB is integrated as an auxiliary to extract modality-shared features. The structure of our \PEB is shown in Fig.~\ref{fig:framework}. Specifically, a convolutional layer (\textit{``Conv1''}) and a deconvolutional layer (\textit{``DConv2''}) are employed to extract high-level features, \ie $\textbf{F}_{p_1}$ and $\textbf{F}_{p_2}$, and restore the resolution of feature maps to that of ground-truth body keypoint heatmaps, which are denoted as follows,
\begin{equation}
\textbf{F}_{p_1}=Conv1(\textbf{F}_S,\theta^1_P),  \quad 
\textbf{F}_{p_2}=DConv2(\textbf{F}_{p_1},\theta^2_P),
\end{equation}
where $Conv1(*,\theta^1_P)$ and $DConv2(*,\theta^2_P)$ represents a $3\times3$ convolutional layer with parameters $\theta_P^1$ and $\theta_P^2$, respectively. Note that both layers are followed by a ReLU activation function, which are omitted in equations for simplicity.

Subsequently, a \RM \cite{osokin2019global} is used to extract refined body keypoint features and predict body keypoint heatmaps. Specifically, the \RM consists of a U-Shaped Block, three Refine Blocks and two convolutional layers. The U-Shaped Block and Refine Blocks are employed to extract refined features. On top of it, the convolutional layers are applied for the heatmap estimation. Mathematically, given high-level features $\textbf{F}_{p_2}$, refined keypoint features $\textbf{F}_R$ and predicted heatmaps $\hat{\textbf{H}}$ can be respectively obtained by,
\begin{align}
\textbf{F}_R = RM_F\big(&\textbf{F}_{p_2},\theta_{RM}^F\big), \\
\hat{\textbf{H}}= RM_H(&\textbf{F}_R, \theta_{RM}^H),
\label{equ:FR}
\end{align}
where $RM_F(*,\theta_{RM}^F)$ and $RM_H(*, \theta_{RM}^H)$ denote the refined feature extraction stage with parameters $\theta_{RM}^F$, and the heatmap estimation stage with parameters $\theta_{RM}^H$ in the \RM, respectively.

\subsubsection{Body keypoint features transferring}
To exploit body keypoint features derived by \PEB in the \RB, a convolutional layer (\textit{``Conv3''}) and a convolutional block (\textit{``Conv Block6''}) are employed to deal with the mismatch in terms of the resolution and the channel number of feature maps. 

Specifically, the refined keypoint features $\textbf{F}_R$ are firstly downsampled by \textit{``Conv3''} so that the derived $\textbf{F}_R'$ has the identical resolution as $\textbf{F}_{p_1}$, which is denoted as,
\begin{equation}
\textbf{F}_{R'} = Conv3(\textbf{F}_R,\theta^3_P),
\end{equation}
where $Conv3(*,\theta^3_P)$ represents the $3\times3$ convolutional layer with a stride of 2 and parameters $\theta^3_P$. On top of that, \textit{``Conv Block6''} aligns the channel number of feature maps from the \PEB with that from the \RB. Therefore, the final body keypoint features $\textbf{F}_P$ can be derived as follows,
\begin{equation}
\textbf{F}_{P} = ConvB6(\textbf{F}_{R'} + \textbf{F}_{p_1} ,\theta^4_P),
\end{equation}
where $ConvB6(*,\theta^4_P)$ denotes the convolutional block with the parameters $\theta^4_P$, consisting of a $3\times3$ convolutional layer with a stride of 2 and a $1\times1$ convolutional layer. Note that all convolutional layers are followed by a ReLU activation function.

\subsubsection{Body keypoint features integration}
To highlight the body keypoint regions in the features output by the ReID branch, the final keypoint features $\textbf{F}_P$ are employed to generate the body keypoint masks $\textbf{M}$, \ie
\begin{equation}
\textbf{M} = sigmoid(\textbf{F}_{P}).
\end{equation}
The values of the masks are regularized by the sigmoid function to [0, 1], which serve as soft attention maps to refine the identity features from the \RB.

\subsection{\RB}
\subsubsection{Global-level feature extraction}
Apart from the \PEB that aids modality-shared features extraction, a \RB is employed to extract ID-related features. As can be seen from Fig.~\ref{fig:framework}, the \RB mainly consists of 2 convolutional blocks (\textit{``Conv Block4\textasciitilde5''}). Following previous works \cite{ye2019bi,ye2020deep,zhu2020hetero}, \textit{``Conv Block4''}, \textit{``Conv Block5''} follow the structures of the third and fourth res-convolution block (\textit{layer3}, \textit{layer4}) of ResNet50 \cite{he2016deep}, respectively. Mathematically, identity features $\textbf{F}_{id_2}$ are extracted by,
\begin{equation}
\textbf{F}_{id_2}= ConvB5(ConvB4(I_m, \theta^1_{ID}), \theta^2_{ID}), m\in\{RGB,IR\},
\end{equation}
where $ConvB4(*,\theta^1_{ID})$ and $ConvB5(*,\theta^2_{ID})$ denote a convolutional block with the parameters $\theta^1_{ID}$ and $\theta^2_{ID}$, respectively. Then, identity features $\textbf{F}_{id_2}$ are refined by the body keypoint masks $\textbf{M}$ derived by \PEB by performing the element-wise product operation $\odot$ to obtain the final identity features $\textbf{F}_{ID}$, \ie
\begin{equation}
\textbf{F}_{ID} =\textbf{F}_{id_2} \odot \textbf{M}.
\end{equation}

\subsubsection{Local-level feature partition}
Since part features can offer fine-grained information for identity identification, PCB models \cite{sun2018beyond} are exploited in the proposed framework for local feature learning. Following \cite{luo2020dynamic,zhu2020hetero}, convolutional features $\textbf{F}_{ID}$ from \RB are firstly partitioned into $P$ horizontal stripes and then transferred to feature vectors via Global Average Pooling (GAP) before being sent to the corresponding PCB model, which can be formulated as follow,
\begin{equation}
\textbf{F}_{ID_1}, ..., \textbf{F}_{ID_P}=GAP\big(Part(\textbf{F}_{ID})\big),
\end{equation}
where $Part(\cdot)$, $GAP(\cdot)$ denote the horizontal partition and GAP, respectively.

As can be seen from Fig.~\ref{fig:framework}, a PCB model consists of a Fully-Connected (FC) layer and a classifier. The former reduces the dimensions of feature vectors from 2048-dim to 512-dim, and the latter is employed for identity prediction. For the $i$-th PCB model, the fine-grained part features $f_{ID_i}$ are obtained by,
\begin{equation}
\label{equ:pcb}
f_{ID_i}=FC_{ID_i}(\textbf{F}_{ID_i}, \theta_{fc_{id}}^i), \quad i=\{1,...,P\},
\end{equation}
where $FC_{ID_i}(*, \theta_{fc_{id}}^i)$ denotes the FC layer with the parameters $\theta_{fc_{id}}^i$. $P$ is the number of PCB models, which is empirically set as 6 in the paper.

The local feature learning is also performed on convolutional features $\textbf{F}_{P}$ from \PEB to obtain the corresponding fine-grained part features $f_{P_i}$. During inference, fine-grained part features are concatenated for the identity retrieval, \ie
\begin{align}
f_{ID} = [f_{ID_1}, ..., f_{ID_P}]_c&, \label{equ:fid} \quad
f_{P} = [f_{P_1}, ..., f_{P_P}]_c, \\
f_{ALL} &= [f_{ID}, f_P]_c,  \label{equ:fall}
\end{align}
where $[\cdot,...,\cdot]_c$ represents the feature concatenation along the channel dimension.
\begin{figure*}[t]
	\centering
	\setlength{\abovecaptionskip}{0.1cm}
	\includegraphics[width=0.8\textwidth]{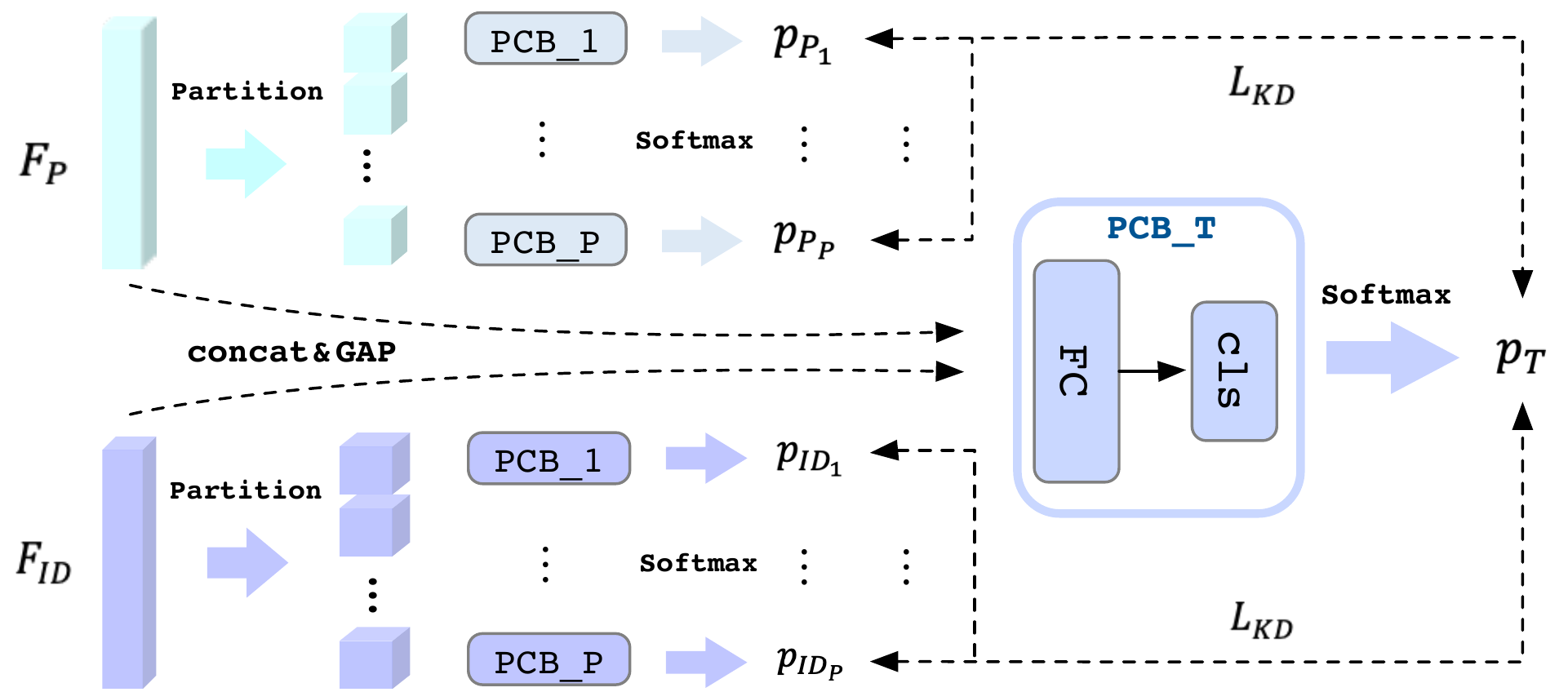}
	\caption{Illustration of \HFC(HFC). GAP represents Global Average Pooling.}
	\label{fig:hfc}
	\vspace{-0.35cm}
\end{figure*}

\subsection{\HFC}
To ensure the discriminability consistency of global and local features, \HFC (HFC) is proposed to bond the learnings of global features and local ones. The structure of HFC is illustrated in Fig.~\ref{fig:hfc}, which is inspired by the Teacher-Student learning spirit in Knowledge Distillation (KD) \cite{hinton2015distilling}. As can be seen, instead of introducing an additional pre-trained teacher model, the predictions of convolutional features are employed as ``soft-targets'' to provide an \textit{extra} supervision for \textit{``Student''} models, \ie PCB models of \PEB and \RB. 

Specifically, convolutional features of \PEB and \RB are firstly concatenated along channel dimension and then transferred to feature vectors by Global Average Pooling. Then, a PCB model (\textit{``PCB\_T''}) is employed to obtain ``soft-targets'', \ie $P_T=\{p_T^i\}_{i=1}^N$. $N$ refers to the number of training images. Formally, given an image $I_i$ with the identity label $y_i$, $p_T^i$ can be obtained as follows,
\begin{equation}
p_T^i =p(y_T^i=y_i|I_i) = \frac{\exp(y_T^i)}{\sum_{k=1}^{N_{id}}\exp(y_T^k)},
\end{equation}
where $y_T^k=w_{t_k}^T f_T^i$, $f_T^i$ is the fine-grained feature of the $i$-th image output by the FC layer in \textit{``PCB\_T''}. $w_{t_k}$ indicates the parameter of the classifier in \textit{``PCB\_T''} for the $k$-th identity. $N_{id}$ is the number of identities in the whole training set. For the $j$-th PCB model of \PEB and \RB, the corresponding probability predictions, \ie $P_{P_j}=\{p_{P_j}^i\}_{i=1}^N$ and $P_{ID_j}=\{p_{ID_j}^i\}_{i=1}^N$, can be calculated in the same way, respectively.

In order to supervise the local feature learning with the global one, KD loss $L_{KD}$ is employed to reduce the distance between two prediction distributions, \ie $P_T$ and $P_{ID(P)_j}$. Given a mini-batch with $M$ images, $L_{KD}$ is formulated as follows, 
\begin{equation}
L_{KD} = \frac{1}{M} \sum_{i=1}^{M} \sum_{j=1}^{P} \big(KL(p_T^i, p_{ID_j}^i) + KL(p_T^i, p_{P_j}^i)\big),
\end{equation}
where $KL(p,q)$ measures the Kullback-Leibler divergence between distribution $p$ and distribution $q$. $P$ denotes the number of PCB models of each branch. 

\subsection{Loss functions}
{\bf Batch sampling method} \quad
Following \cite{zhu2020hetero,luo2020dynamic}, an online batch sampling strategy is adopted during training. Specifically, $D$ identities are randomly selected at each iteration. For each identity, $K$ RGB images and $K$ IR images are then randomly selected to build the mini-batch. Therefore, the batch size $M=2*DK$. In the paper, we set $D=8$, $K=4$, and $M=64$ during training.

{\bf Pose estimation loss}\quad
To encourage \PEB to learn modality-shared features, pose estimation loss $L_{pose}$ is introduced to minimize pixel-wise Euclidean distances between ground-truth body keypoint heatmaps and the predicted ones. In the paper, ground-truth heatmaps are derived by a pose estimation model \cite{osokin2019global}, which is pre-trained on the LIP dataset \cite{liang2018look}. Formally, $L_{pose}$ across a mini-batch is defined by,
\begin{equation}
L_{pose} =\frac{1}{M} \sum_{i=1}^{M}\sum_{x,y}(H_i(x,y) - \hat{H_i}(x,y))^2, 
\label{equ:L_pose}
\end{equation}
where $H_i(x,y)$, $\hat{H_i}(x,y)$ represent the pixel value at $(x,y)$ position of the $i$-th ground truth and the predicted body keypoint heatmap, respectively.

{\bf Identity loss} \quad
To extract ID-related features, identity loss is performed on each PCB model of both \PEB and \RB. Given the probability predictions $P_{P_j}$ and $P_{ID_j}$ given by the $j$-th PCB model of \PEB and \RB, respectively, identity loss $L_{id}$ across a mini-batch is formulated as,
\begin{equation}
L_{id}= -\frac{1}{M} \sum_{i=1}^{M} \sum_{j=1}^{P} (\log p^i_{P_j} + \log p^i_{ID_j} ),
\label{equ:L_id}
\end{equation}
where $P$ denotes the number of PCB models of each branch. 

{\bf Hetero-center triplet loss}\quad
To reduce both intra- and inter-modality discrepancies, hetero-center triplet (HC-tri) loss \cite{liu2020parameter} is also employed for local feature learning. Similar to identity loss, HC-tri loss is also performed on each PCB model. Specifically, for each PCB model, the centers of fine-grained part features of the $i$-th identity in a given mini-batch from RGB modality $c_{i,j}^{RGB}$ and IR modality $c_{i,j}^{IR}$ can be computed as follows,
\begin{equation}
c_{i,j}^{RGB} =\frac{1}{K} \sum_{j=1}^{K} f_{i,j}^{RGB}, \quad
c_{i,j}^{IR} =\frac{1}{K} \sum_{j=1}^{K} f_{i,j}^{IR}, \quad  i = \{1,...,D\},
\end{equation}
where $f_{i,j}^{RGB(IR)}$ represents the fine-grained part feature of the $j$-th RGB (IR) image in the given mini-batch. Therefore, HC-tri loss for the $z$-th PCB model of \PEB is computed as follows,
\begin{small}
\begin{equation}
\begin{split}
L_{hctri}^{P_z} = &\sum_{i=1}^{D}[\rho+||c_{i}^{RGB}-c_{i}^{IR}||_2-\mathop{\rm{min}}_{
	m\in\{RGB,IR\}
	\atop
	j\neq i }||c_{i}^{RGB}-c_{j}^{m}||_2]_+ \\
+&\sum_{i=1}^{D}[\rho+||c_{i}^{IR}-c_{i}^{RGB}||_2-\mathop{\rm{min}}_{
	m\in\{RGB,IR\}
	\atop
	j\neq i }||c_{i}^{IR}-c_{j}^{m}||_2]_+,
\label{equ:L_hc}
\end{split}
\end{equation}
\end{small}
where $[x]_+= max(x, 0)$. $\rho$ refers to the margin value, which is empirically set as 0.3. HC-tri loss of the $z-th$ PCB model of \RB, \ie $L_{hctri}^{ID_z}$, can be calculated in the same way. Therefore, the overall HC-tri loss across the given mini-batch can be derived as follows,
\begin{equation}
L_{hctri} = \sum_{j=1}^{P}(L_{hctri}^{P_j} + L_{hctri}^{ID_j}).
\end{equation}
The overall objective for training is defined as, 
\begin{equation}
L = L_{id} + \beta L_{hctri} + \lambda L_{pose} + \gamma L_{KD},
\end{equation}
where $\beta$, $\lambda$, and $\gamma$ are the weighting factors to balance each loss term, which are empirically set as 0.1, 5, 1 in the paper, respectively.
\linespread{1.2} 
\begin{table*}[htb]
	\caption{Comparison with the state-of-the-art methods on the \textbf{SYSU-MM01} dataset. Rank-r ($r=1,10,20$) accuracy($\%$) and mAP($\%$) are reported. $Ours_{ID}$ and $Ours_{ALL}$ denote the features used for evaluation are obtained from ReID branch and both branches, respectively.}
	\label{tab:sysu}
	\centering
	\setlength{\tabcolsep}{1.5mm}{  
		\begin{tabular}{c|c|cccc||cccc}
			\hline
			\multirow{2}{*}{\centering Method} & \multirow{2}{*}{\centering Venue} &\multicolumn{4}{c}{All-search}  &\multicolumn{4}{c}{Indoor-search} \\
			\cline{3-10}
			&  & Rank-1 & Rank-10 & Rank-20 & mAP & Rank-1 & Rank-10 & Rank-20 & mAP  \\
			\hline
			Zero-Pad \cite{wu2017rgb} &ICCV2017  &14.80  &54.12  &71.33  &15.95 & 20.58  &68.38  &85.79  &26.92 \\
			HCML \cite{ye2018hierarchical} &AAAI2018 &14.32& 53.16& 69.17& 16.16 &24.52 &73.25 &86.73 &30.08 \\
			cmGAN \cite{dai2018cross} & IJCAI2018 &26.97 &67.51 &80.56 &31.49 &31.63 &77.23 &89.18& 42.19 \\
			eDBTR \cite{ye2019bi}& TIFS2019 &27.82& 67.34& 81.34& 28.42 &32.46& 77.42 &89.62& 42.46 \\
			HSME \cite{hao2019hsme} & AAAI2019 & 20.68 &32.74 &77.95 &23.12& - & - & -& - \\
			D$^2$RL \cite{wang2019learning} & CVPR2019 &28.90 &70.60 &82.40 &29.20 &- &-& -& - \\
			MSR \cite{feng2019learning}& TIP2019 & 37.35 &83.40& 93.34& 38.11 &39.64& 89.29 &97.66 &50.88 \\
			AlignGAN \cite{wang2019rgb} & ICCV2019 & 42.40 &85.00 &93.70 &40.70 &45.90 &87.60 &94.40& 54.30 \\
			TSLFN \cite{zhu2020hetero}& Neuro2020 & 56.96 &91.50& 96.82& 54.95 & 59.74 &92.07& 96.22 &64.91 \\
			AGW \cite{ye2020deep} & Arxiv2020 & 47.50 & - & - & 47.65 & 54.17 & - & - & 62.97 \\
			X-Modal \cite{li2020infrared} & AAAI2020& 49.92& 89.79 &95.96& 50.73 &-& -& - &- \\
			MACE \cite{ye2020cross} & TIP2020 &51.64 &87.25& 94.44 &50.11 &57.35 &93.02 &97.47 &64.79 \\
			DDAG \cite{luo2020dynamic} & ECCV2020 &54.75 &90.36 &95.81& 53.02& 61.02 &94.06 &98.41 &67.98 \\
			cm-SSFT \cite{lu2020cross} & CVPR2020 &61.60 &89.20 &93.90 &63.20 &70.50 & 94.90& 97.70 &72.60 \\
			NFS \cite{chen2021neural} & CVPR2021 & 56.91 & 91.34 & 96.52 & 55.45 & 62.69 & 96.53 & 99.07& 69.79\\
			CICL \cite{zhao2021joint} & AAAI2021 & 57.2 & 94.3& 98.4& 59.3& 66.6& 98.8& 99.7& 74.7\\
			GLMC \cite{zhang2021global} & TNNLS2021 & 64.37 & 93.90 & 97.53 & 63.43 & 67.35 & 98.10 & 99.77 & 74.02\\
			LbA \cite{park2021learning} & ICCV2021 & 55.41 & - & - & 54.14 & 58.46 & - & - & 66.33\\
			\hline
			$Ours_{ID}$ & - & 65.82 & 94.53 & 98.23 & 64.52 & 71.74 & 94.57 & 97.60 & 74.54\\
			$Ours_{ALL}$ & - &\textbf{71.21} & \textbf{95.35} &\textbf{98.81} &\textbf{67.15} & \textbf{72.55}& \textbf{97.15} & \textbf{98.60} & \textbf{77.05} \\
			\hline
		\end{tabular}}
\end{table*}
\section{Experiments}
\label{4}
\subsection{Datasets, evaluation metrics and implementation details}
\label{4.1}
{\bf Datasets}\quad
Two benchmark datasets (SYSU-MM01 \cite{wu2017rgb} and RegDB \cite{nguyen2017person}) are employed to evaluate the performance of the proposed method. 

\textbf{SYSU-MM01} \cite{wu2017rgb} consists of images captured by 6 cameras, including 2 IR cameras and 4 RGB ones (2 outdoors and 2 indoors). The training set contains 395 persons, with 22,258 RGB images and 11,909 IR images. The test set contains 96 persons, with 3,803 IR images for query and 301 randomly selected RGB images as the gallery. Following \cite{wu2017rgb}, two evaluation modes are conducted: \textit{All-search} and \textit{Indoor-search}. For \textit{Indoor-search} mode, images collected by indoor RGB cameras are exclusively selected to built the gallery set. For \textit{All-search} mode, images are randomly selected from all RGB cameras to form the gallery set.

\textbf{RegDB} \cite{nguyen2017person} contains 412 identities, with 206 for training and 206 for testing. Each identity has 10 RGB and 10 IR images. Two evaluation modes are employed: \textit{Visible-Thermal} and \textit{Thermal-Visible}. The former refers to searching for corresponding IR images with a RGB image and vice versa. The dataset is randomly split into 10 training/testing trials. The evaluation results are given by averaging the performances over the 10 trials.

{\bf Evaluation metrics}\quad
Following the standard evaluation protocol given by \cite{ye2018hierarchical,zhu2020hetero,ye2020deep}, Cumulative Matching Characteristics (CMC) curve and mean Average Precision (mAP) are adopted as evaluation metrics. Here, CMC reports the probabilities of the targeted identity occurring at top-r in the ranking list, \ie ``Rank-r'' accuracy. mAP measures the overall retrieval performance when multiple matching cases occur in the gallery set.

{\bf Implementation details}\quad
The experiments are deployed on an NVIDIA GeForce 2080Ti GPU with Pytorch. Following most existing works \cite{luo2020dynamic,zhu2020hetero,ye2020deep}, all input images are resized to 288$\times$144. Random cropping, random erasing, and horizontal flipping are adopted for data augmentation. The parameters of \MSP, \MSH, and \RB are initialized by ResNet50 \cite{he2016deep} pre-trained on ImageNet. Other parameters are initialized by Xavier initialization \cite{glorot2010understanding}. We adopt the SGD optimizer with a weight decay of 0.0005 for optimization. The learning rate is initialized as 0.01 and decays by 0.5 at every 20 epoch. The training process iterates for 100 epochs in total.

\subsection{Comparison with state-of-the-arts}
\label{4.3}
We extensively compare our algorithm with the current State-Of-The-Art (SOTA) methods on both SYSU-MM01 \cite{wu2017rgb} and RegDB \cite{nguyen2017person} datasets. The SOTA methods include pioneering ones (Zero-Pad \cite{wu2017rgb} and HCML \cite{ye2018hierarchical}), GAN-based ones (cmGAN \cite{dai2018cross}, AlignGAN \cite{wang2019rgb}, and D$^2$RL \cite{wang2019learning}), middle modality based ones (X-Modal \cite{li2020infrared} and cm-SSFT \cite{lu2020cross}), feature constraints based ones (eBDTR \cite{ye2019bi}, HSME \cite{hao2019hsme}, MSR \cite{feng2019learning}, TSLFN \cite{zhu2020hetero}, AGW \cite{ye2020deep}, and GLMC \cite{zhang2021global}), dual-level feature alignment based ones (DDAG \cite{luo2020dynamic}, MACE \cite{ye2020cross}, CICL \cite{zhao2021joint}, and LbA \cite{park2021learning}).

\textbf{Evaluations on SYSU-MM01}
Table~\ref{tab:sysu} reports the performance of our model and State-Of-The-Art (SOTA) approaches on the SYSU-MM01 \cite{wu2017rgb} dataset. ``$Ours_{ID}$'' and ``$Ours_{ALL}$'' refer to the retrieval performance with features $f_{ID}$ (Equ.~\ref{equ:fid}) and $f_{ALL}$ (Equ.~\ref{equ:fall}), respectively. It can be seen that both ``$Ours_{ID}$'' and ``$Ours_{ALL}$'' outperform SOTA approaches on ALL evaluation metrics in both \textit{All-search} and \textit{Indoor-search} evaluation modes. Compared to the second-best method (GLMC \cite{zhang2021global}), the model performance is improved by approximately 7$\%$ and 4$\%$ with $f_{ALL}$ in terms of the rank-1 accuracy and the mAP score, respectively.
\linespread{1.2} 
\begin{table*}[htb]
	\caption{Comparison with the state-of-the-art methods on \textbf{RegDB} dataset. Rank-r ($r=1,10,20$) accuracy($\%$) and mAP($\%$) are reported. $Ours_{ID}$ and $Ours_{ALL}$ denote the features used for evaluation are obtained from ReID branch and both branches, respectively.}
	\label{tab:regdb}
	\centering
	\setlength{\tabcolsep}{1.5mm}{  
		\begin{tabular}{c|cccc||cccc}
			\hline
			\multirow{2}{*}{\centering Method}&\multicolumn{4}{c}{Visible-Thermal} & \multicolumn{4}{c}{Thermal-Visible} \\
			\cline{2-9}
			& Rank-1 & Rank-10 & Rank-20 & mAP & Rank-1 & Rank-10 & Rank-20 & mAP \\
			\hline
			Zero-Pad \cite{wu2017rgb}  & 17.75 &34.21& 44.35 &18.90 & 16.63 &34.68 &44.25 &17.82 \\
			HCML \cite{ye2018hierarchical} &24.44 &47.53 &56.78 &20.80 & 21.70& 45.02 &55.58 &22.24\\
			eDBTR \cite{ye2019bi}&34.62 &58.96& 68.72& 33.46 &34.21 &58.74 &68.64 &32.49\\
			HSME \cite{hao2019hsme} &  50.85 & 73.36 & 81.66 & 47.00 & 50.15 &72.40& 81.07 &46.16\\
			D$^2$RL \cite{wang2019learning} & 43.40 & 66.10 & 76.30 & 44.10 & - & - & - & -\\
			MSR \cite{feng2019learning}&  48.43 &70.32& 79.95 &48.67 & - & - & - & - \\
			AlignGAN \cite{wang2019rgb} & 57.9 & - & - & 53.6 & 56.3 & - & -& 53.4\\
			X-Modal \cite{li2020infrared} & 62.21 &83.13 &91.72 &60.18 & - & - & - & -\\
			DDAG \cite{luo2020dynamic} & 69.34 &86.19 &91.49& 63.46 & 68.06 & 85.15 & 90.31 & 61.80\\
			AGW \cite{ye2020deep} &  70.05 & - & - & 66.37 & - & - & - & -\\
			MACE \cite{ye2020cross} & 72.37 & 88.40 & 93.59 & 69.09 & 72.12 & 88.07&93.07& 68.57\\
			cm-SSFT \cite{lu2020cross} & 72.3 & - & - & 72.9& 71.0 & - & - & 71.7 \\
			NFS \cite{chen2021neural} & 80.54 & 91.96 & 95.07 & 72.10 & 77.95 & 90.45 & 93.62 & 69.79\\
			CICL \cite{zhao2021joint} & 78.8 & - & - & 69.4 & 77.9 & - & - & 69.4\\
			GLMC \cite{zhang2021global} & 91.84 & 97.86 & 98.98 & 81.42 & 91.12 & 97.86 & 98.69 & 81.06\\
			LbA \cite{park2021learning} & 74.17 & - & - & 67.64 & 72.43 & - & - & 65.46 \\
			\hline
			$Ours_{ID}$ & 92.14 & 98.16 & 99.22 & 87.88 & 91.36 & 97.57 &98.88 & 86.70  \\
			$Ours_{ALL}$ &\textbf{93.35} &\textbf{98.61} &\textbf{99.42} &\textbf{88.98} & \textbf{92.72} & \textbf{98.79} & \textbf{99.36} & \textbf{87.83} \\
			\hline
		\end{tabular}}
\end{table*}

\textbf{Evaluations on RegDB}
The evaluation results on RegDB \cite{nguyen2017person} are shown in Table~\ref{tab:regdb}. It can be observed that the proposed model obtains surprisingly good results in both \textit{``Visible-Thermal''} and \textit{``Thermal-Visible''} modes. Specifically, the performance of $Ours_{ID}$ exceeds GLMC \cite{zhang2021global} by 6.46$\%$ and 5.64$\%$ in terms of the mAP score in two evaluation modes, respectively. The improvements achieve 7.56$\%$ and 6.77$\%$ when $Ours_{ALL}$ is employed for identity retrieval. The comparison results also prove that, with the proposed \HFC (HFC), each feature stripe is embedded with modality-shared and ID-related information. Such advantage leads to a satisfactory identity accuracy even when features with less dimensions are used for identity retrieval.

\subsection{Comparison with pose-guided single-modality person ReID}
\label{sec:CRR}

As aforementioned, employing pose information in the VI-ReID task is extremely challenging due to the massive gap between visible images (source domain) and infrared images (target domain). Although there are some pose-guided single-modality person ReID works, how to enable pose information to facilitate the VI-ReID task has been unrevealed. In the section, several experiments are conducted on the SYSU-MM01 \cite{wu2017rgb} dataset to explore the potential of pose estimation in the VI-ReID task.

\linespread{1.2} 
\begin{table}[b]
	\caption{Comparison with SOTA pose-guided single-modality person ReID methods on the SYSU-MM01 \cite{wu2017rgb} dataset in terms of Rank-1 accuracy($\%$) and mAP($\%$) at \textit{All-search} setting. The dimension of features used for the evaluation is set as 2560 for all experiments. $Ours_{ID}$* means P in Equ.~(\ref{equ:pcb}) is set as 5. Impl. means experiments are implemented with the official source code provided.}
	\label{tab:crr}
	\centering
	\setlength{\tabcolsep}{1.8mm}{
		\begin{tabular}{c|c|c}
			\hline
			Methods & Rank-1 & mAP \\
			\hline
			PGFA(Impl.) \cite{miao2019pose} & 10.04 & 11.45 \\
			PABR(Impl.) \cite{suh2018part} & 40.73 & 42.19 \\
			$Ours_{ID}$* & 62.82 & 61.90 \\
			\hline
		\end{tabular}}
\end{table}
Firstly, the VI-ReID task is considered as the single-modality person ReID task with two different types of images. For each modality, a modality-specific pose-assisted feature extractor is trained. During the inference, features of query set (IR images) and gallery set (RGB images) are derived by the corresponding feature extractor to conduct the cross-modality person retrieval. Taking a state-of-the-art (SOTA) pose-guided single-modality person ReID work, \ie PGFA \cite{miao2019pose}, as an example, where human landmarks obtained by an off-the-shelf pose estimator are rigidly used to generate attention maps to highlight body regions, the performance is reported in the first row of Table~\ref{tab:crr}. As can be seen, although the same pose estimator is used for two modalities, retrieving identities across modalities with modality-specific pose-assisted features can only achieve 11.45$\%$ in terms of the mAP score. The poor performance demonstrates that, the off-the-shelf pose estimators cannot handle the huge gap between the data distribution of two modalities. In this case, the pose information cannot serve as effective modality-shared cues in the VI-ReID task.

\begin{figure}[]
	\centering
	\setlength{\abovecaptionskip}{0.2cm}
	\subfigure[PABR]{
		\begin{minipage}[b]{0.4\textwidth}
			\includegraphics[width=1\textwidth]{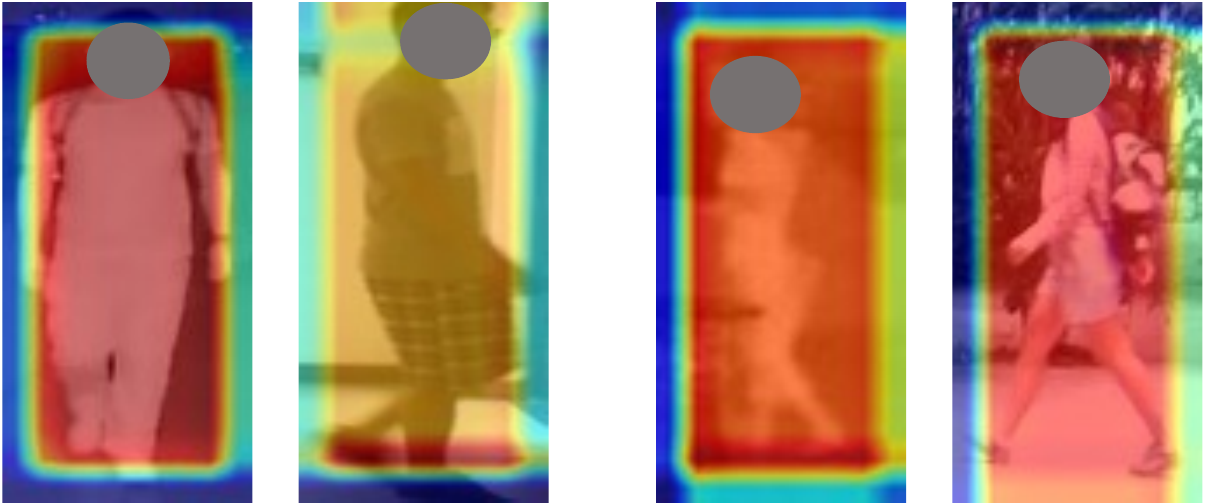}
		\end{minipage}
		\label{fig:PABR}
	}\hspace{8mm}
	\subfigure[$Ours_{ID}$*]{
		\begin{minipage}[b]{0.4\textwidth}
			\includegraphics[width=1\textwidth]{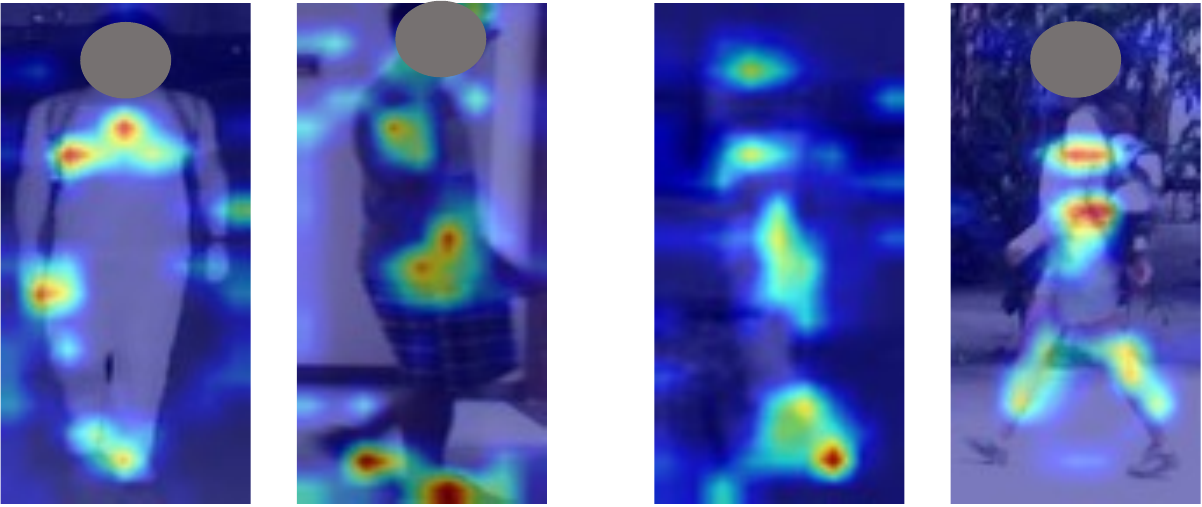}
		\end{minipage}
		\label{fig:ours}
	}
	\caption{Visualization of attentive feature maps output by the pose branch of PABR \cite{suh2018part} and the proposed method on the SYSU-MM01 dataset. For each person (group), the RGB image is shown on the left and the IR image is on the right.}
	\label{fig:biatt}
\end{figure}

To make the best of the pose information in the VI-ReID task, we applied the two-stream network proposed by a single-modality person ReID approach, \ie PABR \cite{suh2018part}, where the ReID branch and the pose branch extract appearance and pose features, respectively. Then two kinds of features are fused for the identity retrieval. To adapt PABR to the VI-ReID task, the cross-modality hard triplet loss \cite{zhao2019hpiln} is used to replace the single-modality one, where the hardest cross-modality triplets are also considered. The performance of PABR in the VI-ReID task is shown in the second row of Table~\ref{tab:crr}. Compared to PGFA, PABR achieves a higher Rank-1 accuracy (40.73$\%$) and mAP score (42.19$\%$). In addition to numerical results, we also visualize the attentive feature maps output by the pose branch of PABR \cite{suh2018part} by means of Grad-CAM \cite{selvaraju2017grad} in Fig.~\ref{fig:PABR}, to have a better understanding of where pose features are extracted. As can be seen, due to the lack of adequate guidance on the pose branch during training, the extracted pose features only locate the whole body coarsely.

To solve the problem, as shown in Fig.~\ref{fig:framework}, the body keypoints generated by a pre-trained pose estimator serve only as the guidance of \PEB in our method. By applying the \textit{pose estimation loss} during training, more fine-grained pose features can be obtained. The performance of the proposed method is shown in the third row of Table~\ref{tab:crr}. It can be seen that, with above improvements, our method outperforms PGFA and PABR by a large margin in terms of the Rank-1 accuracy (62.92$\%$) and the mAP score (61.90$\%$). Additionally, the visualization of the attentive feature maps output by our \PEB is shown Fig.~\ref{fig:ours}. In comparison to PABR, our method focuses on more body details, such as shoulders and feet, which can serve as distinctive cues for the VI-ReID task.

\subsection{Ablation studies}
\label{4.4}
An ablation analysis is conducted on SYSU-MM01 \cite{wu2017rgb} and RegDB \cite{nguyen2017person} to verify the effectiveness of the proposed components in our model. Firstly, a ``Baseline'' model is trained under the supervision of identity loss $L_{id}$ and HC-tri loss $L_{hctri}$, which only consists of \MSP, \MSH, and \RB. The evaluation results are shown in the first row in Table~\ref{tab:abl1}. Based on ``Baseline'', \PEB(PEB), pose estimation loss ($L_{pose}$), and \HFC(HFC) are gradually applied, which results are illustrated in the following rows in Table~\ref{tab:abl1}. Note that the reported results are in the \textit{All-search} mode for the SYSU-MM01 dataset while in the \textit{``Visible-Thermal''} mode for the RegDB dataset.

\linespread{1.2} 
\begin{table}
	\caption{Ablation studies on SYSU-MM01 \cite{wu2017rgb} and RegDB \cite{nguyen2017person} datasets. ``PEB'', ``HFC'' refer to \PEB, and \HFC, respectively.}
	\label{tab:abl1}
	\centering
	\setlength{\tabcolsep}{1.8mm}{
		\begin{tabular}{cccc|c|c|c|c}
			\hline
			&\multicolumn{3}{c|}{Components} &\multicolumn{2}{c|}{SYSU-MM01} & \multicolumn{2}{c}{RegDB} \\
			\cline{2-8}
			& PEB & $L_{pose}$ & HFC & Rank-1 & mAP & Rank-1 & mAP \\
			\hline
			Baseline&&&&57.03 & 56.21 & 85.76 & 77.21\\
			\hline
			\multirow{3}{*}{$Ours_{ID}$}& \checkmark &&& 60.19 & 58.96 & 87.39 & 81.75 \\
			&\checkmark &\checkmark&& 63.90 & 62.79 & 89.25 & 84.16 \\
			&\checkmark &\checkmark&\checkmark&\textbf{65.82} & \textbf{64.52}& \textbf{92.14}& \textbf{87.88} \\
			\hline
			\multirow{3}{*}{$Ours_{ALL}$} &\checkmark &&& 65.05 & 62.78 & 89.53 & 84.12 \\
			&\checkmark &\checkmark && 68.84 & 65.94 & 91.96 & 87.08\\
			&\checkmark &\checkmark&\checkmark&\textbf{71.21} & \textbf{67.15} & \textbf{93.35} & \textbf{88.98} \\
			\hline
		\end{tabular}}
\end{table}

\textbf{Effectiveness of \PEB (PEB)}
As can be seen from the 2\textit{nd} row in Table~\ref{tab:abl1}, by integrating the PEB branch that is pre-trained on a pose estimation dataset, the framework yields an increase of approximately 2$\%$ and 4$\%$ in terms of the mAP score on the SYSU-MM01 dataset and the RegDB dataset, respectively, when $f_{ID}$ is applied for identity retrieval. Likewise, the improvements exceed 6.5$\%$ when employing $f_{ALL}$, which clearly demonstrates the effectiveness of our PEB branch.

Apart from $L_{id}$ and $L_{hctri}$, $L_{pose}$ is further employed on such a structure to ensure the body skeleton points are precisely estimated. The results for $f_{ID}$ and $f_{ALL}$ are shown in the 3\textit{rd} and 6\textit{th} rows, respectively. Specifically, an increase of 3.83$\%$ and 2.41$\%$ can be found in terms of the mAP score on two benchmark datasets for $f_{ID}$. Similar increase, \ie 3.16$\%$ and 2.96$\%$, can be obtained if using $f_{ALL}$.

\textbf{Effectiveness of \HFC (HFC)}
It can be observed from the 4\textit{th} row that HFC yields a satisfactory improvement for $f_{ID}$ in terms of the mAP score, which are 1.73$\%$ and 3.72$\%$ on the SYSU-MM01 dataset and the RegDB dataset, respectively. A similar improvement can also be found for $f_{ALL}$ (the 7\textit{th} row), which are around 2$\%$ for both datasets. The boost of performance proves that the proposed HFC encourages the information flow between global and local features by introducing an extra supervision on part-level features.

\begin{figure*}[htb]
	\centering
	\subfigure[]{
		\begin{minipage}[b]{0.3\textwidth}
			\includegraphics[width=1\textwidth]{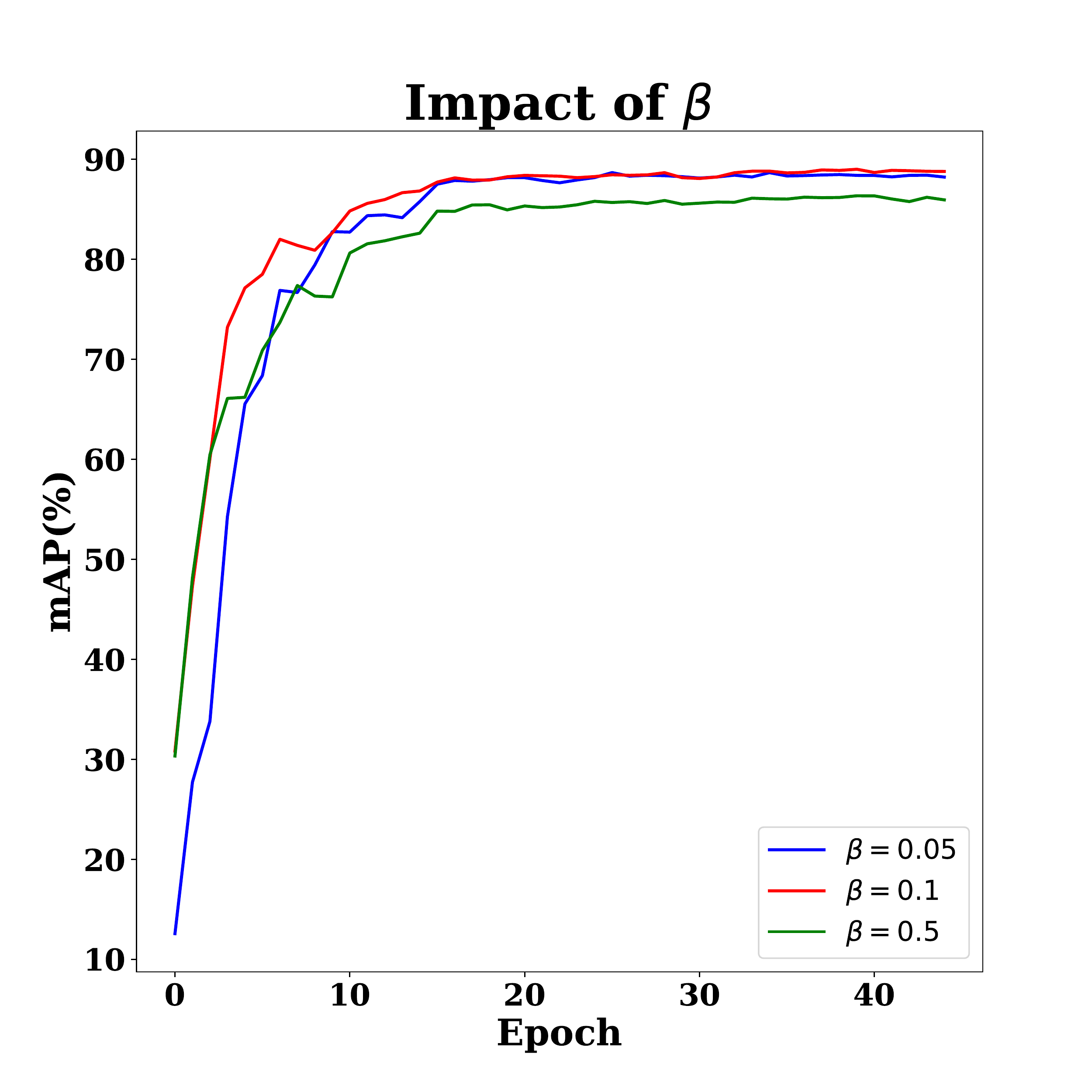}
		\end{minipage}
		\label{fig:beta}
	}\hspace{0mm}
	\subfigure[]{
		\begin{minipage}[b]{0.3\textwidth}
			\includegraphics[width=1\textwidth]{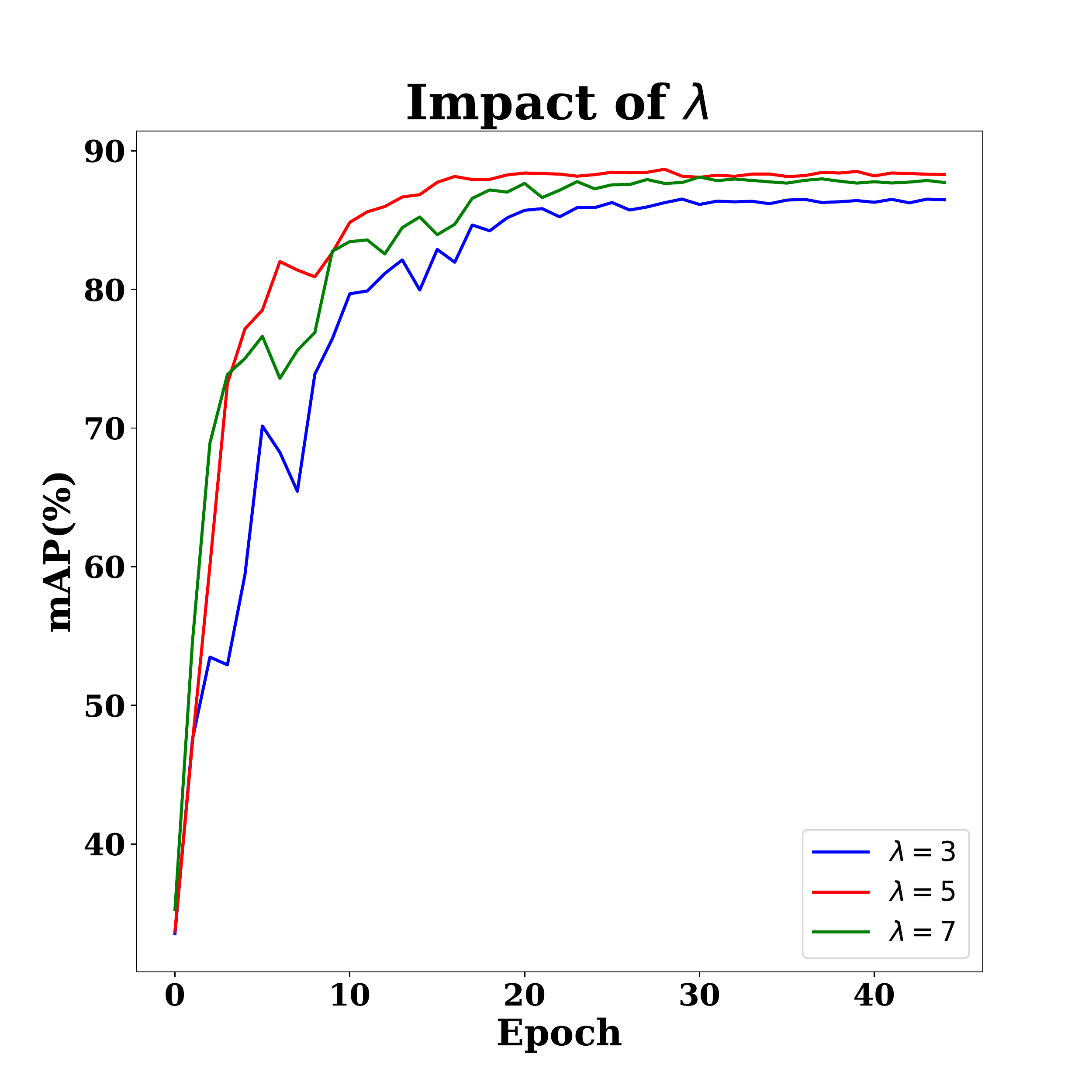}
		\end{minipage}
		\label{fig:lambda}
	}\hspace{0mm}
	\subfigure[]{
		\begin{minipage}[b]{0.3\textwidth}
			\includegraphics[width=1\textwidth]{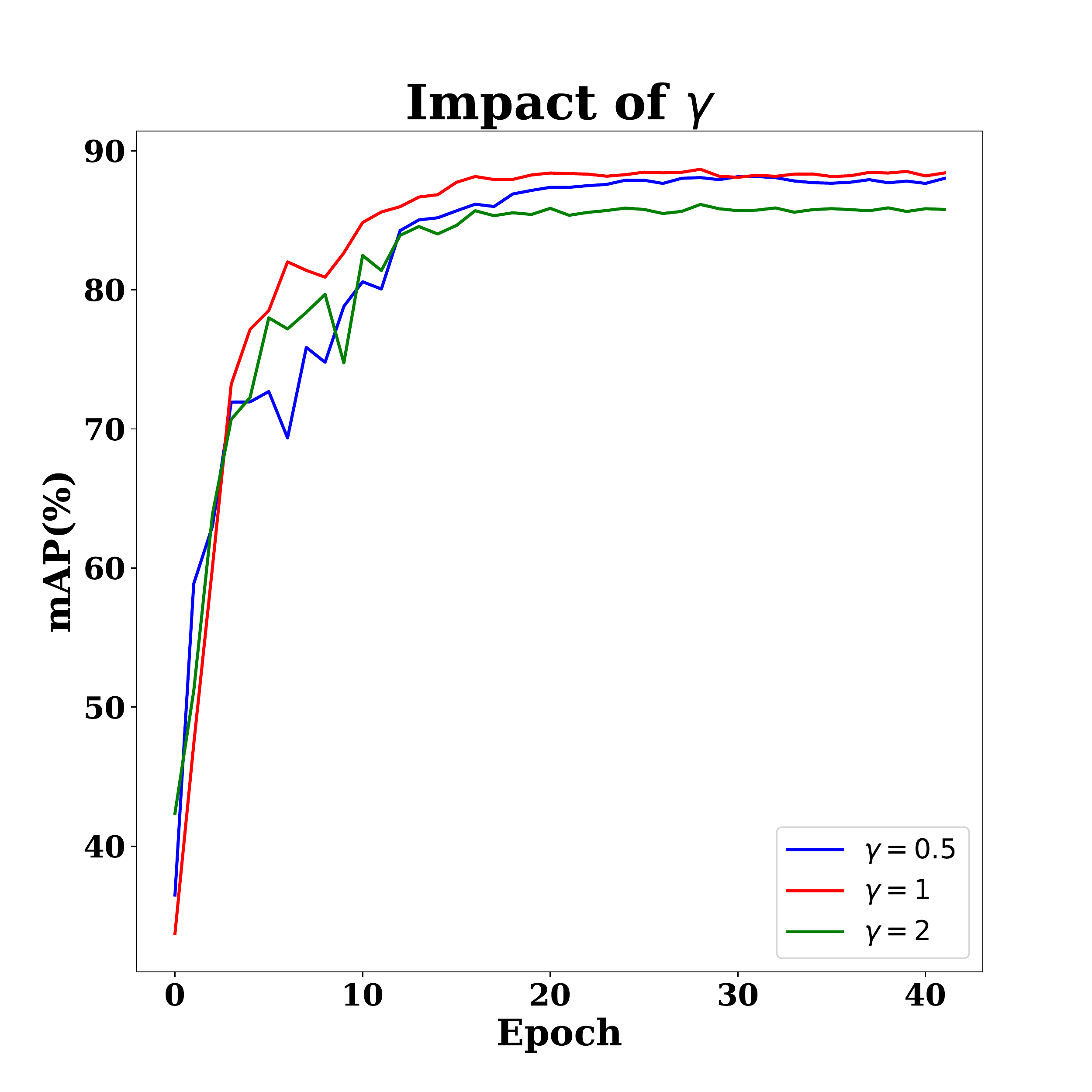}
		\end{minipage}
		\label{fig:gamma}
	}
	\caption{Parameter analysis on the RegDB dataset. \textbf{Best viewed in color.}}
	\label{fig:factor}
\end{figure*}

\subsection{Impact of Loss coefficient}
In this section, we aim to investigate an optimal combination of weight factors to balance multiple loss terms. According to the paper, the overall objective $L$ for training is defined as, 
\begin{equation}
L = L_{id} + \beta L_{hctri} + \lambda L_{pose} + \gamma L_{KD}.
\end{equation}
$\beta$, $\lambda$, and $\gamma$ adjust the contribution of Hetero-center triplet loss, Pose estimation loss, and KD loss, respectively. For a fair comparison, we only change one factor at a time while keeping the rest fixed. Subsequently, models are trained by objectives with different combinations of weight factors. The corresponding mAP scores on the test set are depicted in Fig.~\ref{fig:factor}. 

It can be observed that no matter which combination is employed, the proposed framework could achieve a stable performance within 50 epoch. Additionally, according to the experimental results, the optimal combination $\{\beta,\lambda,\gamma\}=\{0.1,5,1\}$ is chosen during the training.
\begin{figure*}[]
	\centering
	\setlength{\abovecaptionskip}{0.2cm}
	\subfigure[Baseline]{
		\begin{minipage}[b]{0.43\textwidth}
			\includegraphics[width=1\textwidth]{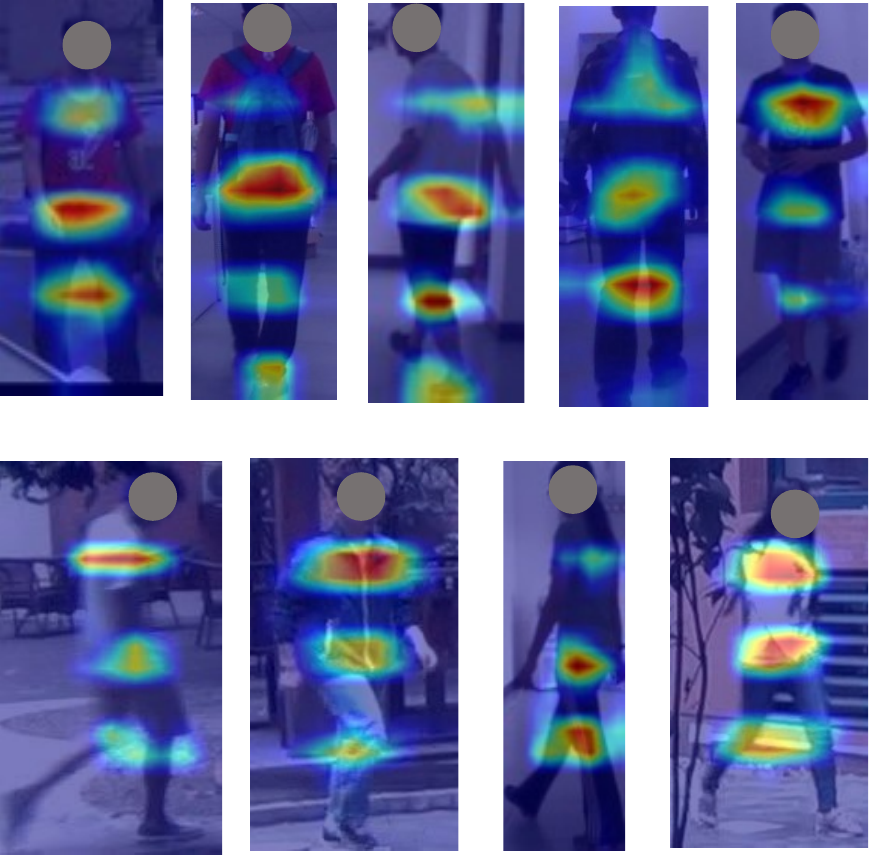}
		\end{minipage}
		\label{fig:hb}
	}\hspace{8mm}
	\subfigure[Ours]{
		\begin{minipage}[b]{0.4\textwidth}
			\includegraphics[width=1\textwidth]{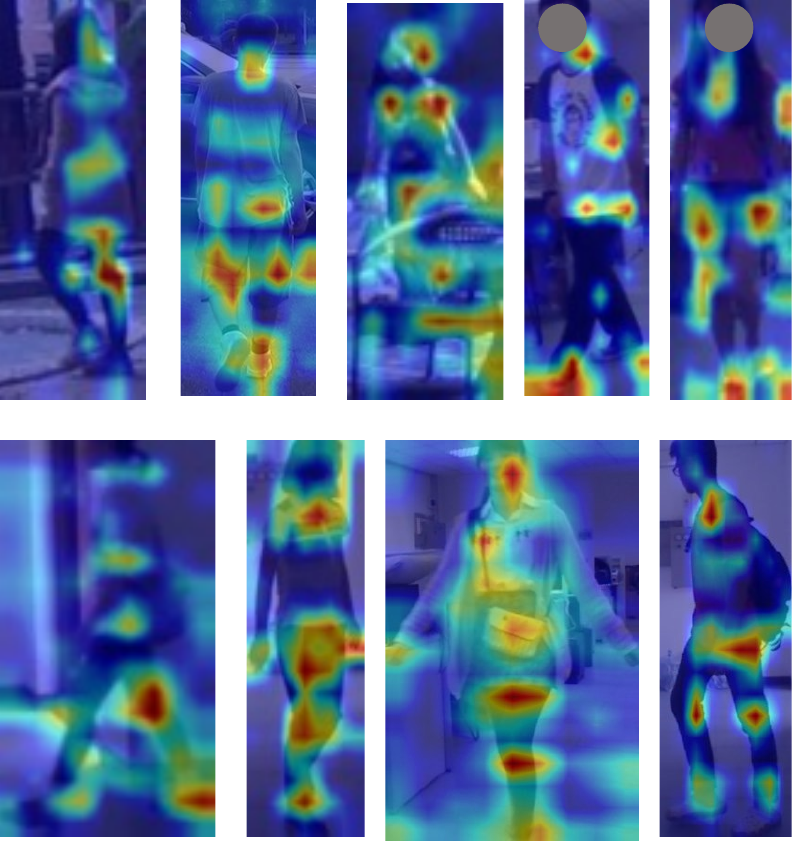}
		\end{minipage}
		\label{fig:hmy}
	}
	\caption{Visualization of gradient feature maps output by \RB on the SYSU-MM01 dataset.}
	\label{fig:heat}
	\vspace{-0.3cm}
\end{figure*}

\subsection{Visualization}
Apart from quantitative results, we also visualize the gradient feature maps output by \RB with Grad-CAM \cite{selvaraju2017grad}, in order to examine where the features are extracted.

Visualization results of ``Baseline'' and the proposed framework (``Ours'') are illustrated in Fig.~\ref{fig:hb} and Fig.~\ref{fig:hmy}, respectively. As can be seen, instead of rigidly extracting features from several vertical regions, the proposed method extracts features from body skeleton joints, which are not only ID-related but highly immune to viewpoint changes and modality changes. The visualization results not only intuitively reveal the reason why our method performs better, but also show the potential of pose estimation task in the field of VI-ReID.

\section{Conclusion}
We have proposed a novel two-stream VI-ReID framework, where modality-shared and ID-related features for identity retrieval are extracted by means of learning an auxiliary task (pose estimation) and the main task (person ReID) simultaneously. By imposing pose estimation and ReID constraints on the \PEB at the same time, both modality-shared and ID-related are fully embedded on each feature stripe. Apart from learning discriminative features at the local level, we also propose a \HFC to bond the learning of global features with local ones by employing the knowledge distillation strategy to ensure the discriminability consistency. The proposed framework achieves a new state-of-the-art on VI-ReID benchmarks in terms of the rank-1 accuracy and the mAP score. 


\bibliographystyle{IEEEtran}
\bibliography{IEEEabrv,PoseReID}
\end{document}